\documentclass[nohyperref]{article}

\usepackage{microtype}
\usepackage{graphicx}
\usepackage{subfigure}
\usepackage{booktabs} 
\newcommand{\usol}{\psi}
\newcommand{\wout}[0]{ {W_\text{out} }}
\usepackage{hyperref}

\usepackage[accepted]{ai4science} 
\usepackage{amsmath}
\usepackage{amssymb}
\usepackage{mathtools}
\usepackage{amsthm}

\usepackage[capitalize,noabbrev]{cleveref}

\theoremstyle{plain}

\theoremstyle{definition}

\theoremstyle{remark}

\usepackage[textsize=tiny]{todonotes}

\icmltitlerunning{One-Shot Transfer Learning of Physics-Informed Neural Networks}

\begin{document}

\twocolumn[
\icmltitle{One-Shot Transfer Learning of Physics-Informed Neural Networks}




\begin{icmlauthorlist}
\icmlauthor{Shaan Desai}{a,b}
\icmlauthor{Marios Mattheakis}{b}
\icmlauthor{Hayden Joy}{b}
\icmlauthor{Pavlos Protopapas}{b}
\icmlauthor{Stephen Roberts}{a}
\end{icmlauthorlist}

\icmlaffiliation{a}{Machine Learning Research Group, University of Oxford, Oxford, United Kingdom}
\icmlaffiliation{b}{School of Engineering and Applied Science, Harvard University, Cambridge, MA, U.S.A.}
\icmlcorrespondingauthor{Shaan Desai}{shaandesai@live.com}

\icmlkeywords{Machine Learning, physics, surrogates, deep learning}

\vskip 0.3in
]



\printAffiliationsAndNotice{} 

\begin{abstract}
Solving differential equations efficiently and accurately sits at the heart of 
progress in many areas of scientific research, from classical dynamical systems to quantum mechanics. There is a surge of interest in using Physics-Informed Neural Networks (PINNs) to tackle such problems as they provide numerous benefits over traditional numerical approaches. Despite their potential benefits for solving differential equations, transfer learning has been under explored. In this study, we present a general framework for transfer learning PINNs that results in one-shot inference for linear systems of both ordinary and partial differential equations. This means that highly accurate solutions to many unknown differential equations can be obtained instantaneously without retraining an entire network. We demonstrate the efficacy of the proposed deep learning approach by solving several real-world problems, such as first- and second-order linear ordinary equations, the Poisson equation, and the time-dependent Schr\"{o}dinger complex-value partial differential equation.
\end{abstract}

\section{Introduction}

Differential equations are used to study a broad array of phenomena, from infection models in biology \cite{kaxiras_first_2020} to chaotic motion in physics \cite{choudhary_physics_2019}. As such, our ability to efficiently and accurately solve these equations under various conditions remains a critical challenge in the scientific community. While traditional approaches such as Runge-Kutta \cite{dormand_families_1987} and Finite Element Methods are well studied and provide solutions of high fidelity, recently, Physics-Informed Neural Networks (PINNs) \cite{raissi_physics-informed_2019,karniadakis_physics-informed_2021} have attracted significant attention as an alternative framework for solving differential equations. PINNs are Neural Networks (NNs) capable of leveraging data and known physical constraints to identify solutions to differential equations. They have numerous advantages over traditional approaches such as being able to easily incorporate data, eliminating the need for a numerical integrator, being able to generate continuous and differentiable solutions, improving accuracy in high dimensions, and maintaining memory efficiency \cite{karniadakis_physics-informed_2021}. However, one of the limitations of using PINNs is the computational expense associated with training networks for different but closely linked tasks. To address this, we explore how transfer learning - the method of training a network on a task and transferring it to new tasks, can be be used to overcome this bottleneck.

Specifically, we show that a PINN, pre-trained on a family of differential equations, can be effectively re-used to solve new differential equations. By freezing the hidden layers of a pre-trained PINN, we demonstrate that solving new differential equations of the same family reduces to optimizing/fine-tuning a linear layer. Furthermore, we specifically show that in the special case of linear systems of differential equations, this optimization is equivalent to solving the normal equations for a  latent space of learnt functions. This implies that the optimal linear weights needed to satisfy a new differential equation can be computed in one-shot with the computational cost of a matrix inversion. This therefore entirely eliminates the need for further training/fine-tuning, dramatically reducing the training overhead by orders of magnitude while maintaining high fidelity solutions. We investigate the efficiency of this approach by solving several ordinary differential equations (ODEs) as well as partial differential equations (PDEs) of practical interest. For many systems, we are able to identify highly accurate solutions to unseen differential equations in a fraction of the time needed to train the equations from scratch.



%


\section{Background}
The general form of an explicit $n^\text{th}$ order ODE can be written as:
\begin{equation}
F(t,\usol,\usol^{(1)},....,\usol^{(n-1)}) = \usol^{(n)},
\label{eqn.ode1}
\end{equation}
where $\usol^{(i)}=\frac{d^i \usol}{dt^i}$ is the $i^\text{th}$ derivative of the solution $\usol(t)$ with respect to the independent time variable $t$. Non-homogeneous linear ODEs, a subclass of the general form of Eqn.~\ref{eqn.ode1}, can be represented as follows:
\begin{equation}
    \hat{D}_n \usol = f(t);~~~ \hat{D}_n \usol= \sum_{i=0}^n a_i(t)\usol^{(i)},
\label{eqn.ode2}
\end{equation}
where $n$ denotes the order of the ODE, $f(t)$ is considered a forcing (or control) term that influences the homogeneity of the solution, and $a_i(t)$ is a time dependent coefficient for each derivative. 


Traditionally, when an ODE is known \textit{apriori}, the differential equation can be solved using integrators (such as Runge-Kutta) for given initial conditions (ICs). Recently, it has been shown that neural networks can be used to efficiently determine accurate solutions to such problems. One such approach uses PINNs. PINNs use a neural network, with weights parametrized by $\theta$, to transform an input $t$ to output solutions $\usol_{\theta}(t)$. Then, by leveraging backpropagation and autograd \cite{maclaurin_autograd_nodate}, exact derivatives of the network output can be computed with respect to the input $\frac{\partial\usol}{\partial t}$. Therefore, given  ICs $u_{\text{ic}} = [\usol_0,\usol_0^{(1)},..,\usol_0^{(n-1)}]^T$, and  known differential operator $\hat{D}_n$ and force $f(t)$, the loss function of a PINN is defined as:
\begin{equation}
    \mathcal{L} = (\hat{D}_n \usol_{\theta}(t) - f(t))^2 + (\bar{D}_0\usol_{\theta}(t)- \usol_{\text{ic}})^2,
    \label{eqn.loss}
 \end{equation}
where $\bar{D}_0\usol = [\usol(0),\usol^{(1)}(0),...,\usol^{(n-1)}(0)]^T$. The first term enforces the differential equation and the second enforces the initial conditions.

Indeed, such a loss can be enforced for other  architectures such as NeuralODE \cite{chen_neural_2018} and Reservoir Computing \cite{rcode_2021} which exploit recurrent neural networks. However, such networks are not as easily adaptable to PDEs as PINNs are. Many well known PDEs such as the diffusion, wave as well as Schr\"odinger equation, can be modeled using PINNs. With PDEs, additional variables are used as inputs $[t,x,y,...]$, so the output is a function of these inputs $\usol(t,x,y,..)$ subject to certain ICs and boundary conditions (BCs). As such, a loss function similar to Eqn. \ref{eqn.loss} can be defined for PDEs. 





The benefits of using a NN architecture to solve differential equations (both ODEs and PDEs) over traditional methods (such as Runge-Kutta or Finite Elements) include rapid inference, elimination of the curse of dimensionality \cite{highdim}, no accumulation of errors as would be found with integrators, continuously differentiable solutions, and low memory cost \cite{karniadakis_physics-informed_2021}. While these benefits have been extensively explored across a range of applications \cite{sirignano_dgm_2018,mattheakis_hamiltonian_2020,zhaii_inferring_2021,wangg_train_2021,mcclenny_self-adaptive_2020,de_wolff_towards_2021}, a limited study exists on how transfer learning techniques can be used \cite{guo_transfer_2021,wangg_train_2021, rcode_2021}. Transfer learning was first developed to accelerate network optimization in computer vision where large datasets are costly to re-train on for specific tasks. The main idea behind the approach is to train a neural network on a large corpus of data and then to freeze the network and re-use some of the layers for new, unseen tasks. This was shown to dramatically reduce training time while maintaining high fidelity solutions in vision. With this in mind, we decided to apply transfer learning to PINNs to significantly accelerate the training on new equations. In doing so, we identify a novel one-shot transfer learning framework for systems of linear differential equations that significantly speeds up inference on unseen linear systems.

\section{Related Work}
Constraining neural networks to learn solutions to differential equations was first introduced by Lagaris et al (\citeyear{lagaris_artificial_1998}). The authors showed that partial derivatives of a neural network output with respect to its inputs can be analytically computed when the architecture of the network is known. Therefore, given the solution and its derivatives,  it is possible to simultaneously enforce the underlying differential equation as well as the ICs and BCs. Indeed this approach forms the basis of PINNs \cite{raissi_physics-informed_2019, karniadakis_physics-informed_2021} where the analytic derivatives from Lagaris et al (\citeyear{lagaris_artificial_1998}) are replaced with backpropagation, i.e. replacing the need  for an analytic derivation of the partial derivatives. PINNs have since been extensively used across many applications including non-linear structures \cite{zhang_physics-informed_2020}, fluid flow on large domains \cite{wangg_train_2021}, moving boundaries \cite{wang_deep_stefan_2021}, inferring micro bubble dynamics \cite{zhaii_inferring_2021}, cardiac activation mapping \cite{sahli_costabal_physics-informed_2020},  ocean modelling \cite{de_wolff_towards_2021}, bundle solvers \cite{flamant_solving_2020} and stochastic and high-dimensional PDEs \cite{karniadakis_physics-informed_2021,yang_physics-informed_2018, sirignano_dgm_2018}. 

Recently, this technique has been extensively used to learn underlying dynamics from data - a concept first proposed by Howse et.al (\citeyear{howse_gradient_1996}). For example, numerous works show that energy conserving trajectories can be effectively learnt from data by enforcing known energy constraints such as Hamiltonians \cite{greydanus_hamiltonian_2019,sanchez-gonzalez_hamiltonian_2019}, Lagrangians \cite{cranmer_lagrangian_2020} and variational integrators \cite{saemundsson_variational_2020,PhysRevE.104.035310} into networks. Extensions in this direction have pushed the envelope to also learn non-conservative/irreversible systems \cite{yin_augmenting_2021,desai_port-hamiltonian_2021,zhong_dissipative_2020,lee_machine_2021} and contact dynamics from sparse, noisy data  \cite{hochlehnert_learning_2021}. 

Other alternative methods to solve equations such as the fourier neural operator \cite{fno} have also emerged. These networks highlight how operator regression coupled with fourier transforms can help us identify the underlying dynamics of physical systems from data.
To increase the pace of innovation across these methods, several software packages have been developed that use neural networks and the backpropagation technique of PINNs to approximate solutions of differential equations such as NeuroDiffEq \cite{neuroDiff2020}, DeepXDE \cite{deepXDE}, and SimNet \cite{simnet2020}.

In spite of these developments, transfer learning remains under explored. Wang et.al (\citeyear{wangg_train_2021}) show transfer learning methods can be used to stitch solutions together to resolve a large domain. Yet further work by Mattheakis et.al (\citeyear{rcode_2021}) illustrates how a reservoir of weights can be transferred to new ICs. Here, we push these further and identify a general model-agnostic method to do one-shot inference for systems of linear ordinary and partial differential equations.

\section{Method}

\subsection{ODEs}

We define a neural network such that the approximate network solution $\usol(t)$ at time points $t$ is: $\usol(t) =H(t)_{\theta_H}W_{\theta_W} + B_{\theta_B}$. In other words, the neural network, parametrized by $\theta=[\theta_H,\theta_W,\theta_B]$, transforms the inputs $t \in \mathbb{R}^{t\times 1}$ into a high dimensional, non-linear latent  space $H \in \mathbb{R}^{t\times h}$ through a composition of non-linear activations and hidden layers. Then, a linear combination of the latent space  is taken, akin to reservoir computing \cite{Jaeger2004}, to obtain the solution $\usol(t)$. 

To train the network, we design the final weights layer to consist of multiple outputs, i.e. $W_{\theta_W} \in \mathbb{R}^{h\times q}$. This is done so that multiple ($q$) solutions, $\usol(t) \in \mathbb{R}^{t\times q}$, can be estimated and simultaneously trained to satisfy equations that have different
linear operators $D_n$ defined by different coefficients $a_i(t)$, as well as different initial conditions $\usol_{\text{ic}}$, and forces $f$. Bundle training allows us to (1) integrate the training into a single network and (2) to encourage the hidden states $H(t)$ to be versatile across equations.

At inference, the weights for the hidden layers are frozen and $H$ is computed at specific time points $\hat{t}$. The solution is therefore $\usol(\hat{t})=H(\hat{t})\wout$ where $\wout$ is trainable. For a new set of ICs  $\usol_{\text{ic}}'$, source $f'$, and differential operator $\hat{D}_n'$ the loss of the linear ODE (Eqn. $\ref{eqn.loss}$), becomes:
\begin{align}
    \mathcal{L} &= \mathcal{L}_{\text{diffeq}} + \mathcal{L}_{\text{IC}} \nonumber \\
    &= \left(\hat{D}_n' H\wout - f'(t)\right)^2 + \left(\bar{D}_0H\wout - \usol_{\text{ic}}'\right)^2
    \label{eqn.loss2}
\end{align}
Since  Eqn. \ref{eqn.loss2}  is convex, the fine-tuning of $\wout$ can be computed analytically. In other words, to minimize $\mathcal{L}$  we need to solve the equation  ${\partial\mathcal{L}}/{\partial\wout}=0$. The derivative of the first term of Eqn. \ref{eqn.loss2} is:
\begin{equation}
    \frac{\partial\mathcal{L}_{\text{diffeq}}}{\partial \wout} = 2\left(\hat{D}_n' H\right)^{T}\left(\hat{D}_n' H \wout- f'(t)\right).
\end{equation}
Taking the same approach for the second term of Eqn. \ref{eqn.loss2} that enforces  ICs, we obtain:
\begin{equation}
    \frac{\partial\mathcal{L}_{\text{ICs}}}{\partial \wout} = 2(\bar{D}_0H)^{T}(\bar{D}_0H \wout - \usol_{\text{\text{ic}}}').
\end{equation}
We let  $\hat{D}_n'H = \hat{D}_{H}$ and $\bar{D}_0H = \bar{D}_H $ to simplify the notation.  Adding the loss terms together and setting them to zero yields the optimal output weights:
\begin{align}
    \wout =   \left( \hat{D}_H^T  \hat{D}_H + \bar{D}_H^T\bar{D}_H\right)^{-1}\left(D_H^Tf'(t) + \bar{D}_H^T\usol_{\text{ic}}'\right).
    \label{eqn.wout_analytic}
\end{align}
Therefore, given any fixed hidden states $H(\hat{t})$ at fixed time-points $\hat{t}$, one can analytically compute a $\wout$ for any linear differential equation that minimizes $\ref{eqn.loss2}$. Broadly, we can think of $H$ as being a  collection of non-orthogonal basis functions that can be linearly combined to determine the output function. 

Note that one special outcome of this formalism is that the matrix inversion at inference is independent of the ICs $\usol_{\text{ic}}'$ and force $f'$, which means for any new ICs or $f'$, $\wout$ can be computed with a simple matrix multiplication if the inverse term in Eqn. \ref{eqn.wout_analytic} is  pre-computed. The benefits of this approach are multi-fold, given $H$ we achieve fast inference (order of seconds for 1000s of differential equations), eliminate the need for gradient-based optimization as no further training is required, and maintain high accuracy if $H$ is well-trained. Indeed this approach relies on determining an inverse matrix. If $D_H^TD_H$ has a large condition number, the matrix will have many eigenvalues close to zero - indicating ill-conditioning. Experimentally, we circumvent this issue by using regularisation or QR decomposition (see Appendix).

We have shown that an analytic $W_{\text{out}}$ can be determined for linear non-homogeneous ODEs. However, the proposed network design can still be used, as we show later, for efficient transfer learning of non-linear ODEs. 

\begin{table*}[t!]
\resizebox{\textwidth}{!}{
    \begin{tabular}{c|c|c|c|c}
    \textbf{Differential Equation} & \textbf{\# Training Bundles} & \textbf{\# Test Bundles} & \textbf{Test Time (s)} & \textbf{Test Accuracy (MSE)} \\
    \hline
    First-order linear  ODEs &10   & 1000  & $7.4 \times 10^{-3}$ & $1.35 \pm 1.65 \times 10^{-10}$ \\
    Second-order linear ODEs &10    & 1000  & $3.4 \times 10^{-3}$ & $2.84 \pm 1.87 \times 10^{-9}$ \\
    Coupled linear oscillators  &10  & 100  & $4.7\times 10^{-2}$ & $2.29 \pm 4.74 \times 10^{-12}$ \\
    Nonlinear oscillator  & 5     & 30    & 5.2 & $1.47 \pm 3.88 \times 10^{-4}$ \\
    Poisson &4 & 100   & 33.2 & $3.60 \pm 8.84 \times 10^{-5}$ \\
    Schr\"{o}dinger & 3     & 400   & 19.4  & $5.02 \pm 8.92 \times 10^{-5}$ \\
    \end{tabular}%
}
\caption{Summary results of our method on all the systems investigated. Training on a few bundles is sufficient to rapidly and accurately scale to many unseen conditions. Note that the nonlinear oscillator is optimized using gradient descent whereas the other methods are all optimized using analytic $\wout$. For reference, training a PINN requires several thousand iterations to obtain accurate solutions, where a single iteration costs ~0.07s. All times are reported for a CPU.}
\label{table:results}%
\end{table*}%

\subsection{PDEs}
An important outcome of the formalism for ODEs is a natural extension to linear PDEs. Many PDEs that appear in real-world problems are linear, including the diffusion equation, Laplace equation, the wave equation as well as the time-dependent Schr\"odinger PDE. 
Considering one spatial dimension $x$ and one time dimension $t$, a general  linear second order differential equation  takes  the form: 
\begin{equation}
    \left(D^t + D^x + D^{xt} + V(t,x)\right)\usol(x,t) = f(x,t),
    \label{eqn:pde}
\end{equation}
where we denote a   second order time  operator  $D^t \usol = \sum_{i=1}^2 a_{i}(t,x) \usol^{(i)}_t$, the spatial second-order operator $D^x \usol = \sum_{i=1}^2 b_{i}(t,x)\usol^{(i)}_x$, and a mixed space-time operator, $D^{xt}\usol = D^{tx}\usol= c(x,t)\usol_{xt}$. The coefficients $a, b, c$ and commonly called  source and  potential $f, V$ functions, respectively,  are  continuous functions of $x,t$, where the lower indices indicate partial derivatives according to the notation: $\usol_\nu^{(i)} = \frac{\partial^{(i)} \usol}{\partial \nu^{(i)}}$ and $\usol_{\nu \nu'}=\frac{\partial^2\usol}{\partial \nu \nu'}$. The structure of Eqn. \ref{eqn:pde} can generalize to higher orders and for more variables.

The last part to complete the derivation is to enforce the BCs and ICs in the loss function. For the purpose of the derivation, we use Dirichlet BCs. Thus,
\begin{align}
    \mathcal{L} &= \mathcal{L}_{\text{diffeq}} + \mathcal{L}_{\text{IC}} + \mathcal{L}_{\text{BCs}} \nonumber \\
    &= \left(\hat{D} \usol - f(t,x)\right)^2 + \left(\usol(0,x)-g(x) \right)^2 \nonumber \\
 & \quad \quad+ \sum_{\mu=L,R}\left(\usol(t,\mu)-B_\mu(t)\right)^2,
\label{eq:loss_pde}
\end{align}
where $\hat{D} = (D^t + D^x + D^{xt}) + V(t,x)$, $B_L(t)$ and $B_R(t)$ are the left and right boundary conditions, and $g(x)$ is the initial condition at $t=0$.
Similarly to the derivation for ODEs, we analytically compute  $\wout$ of  Eqn. \ref{eq:loss_pde}, namely we solve the equation $\partial \mathcal{L}/\partial \wout =0$ considering a neural solution of the form $\psi=H\wout$. Starting with the first term of Eqn. \ref{eq:loss_pde},  we read:
\begin{equation}
    \frac{\partial \mathcal{L}_{\text{diffeq}}}{\partial \wout} 
    = 2\hat{D}_H^T(\hat{D}_H\wout - f(t,x))
\end{equation}
where $\hat{D}_H = \hat{D} H$.
Accordingly, for the IC loss component we obtain:
\begin{equation}
    \frac{\partial \mathcal{L}_{\text{\text{IC}}}}{\partial \wout} = 2H_0^T(H_0\wout - g(0,x)),
\end{equation}
with $H_0=H(0,x)$. For the BCs loss components we have:
\begin{equation}
    \frac{\partial \mathcal{L}_{\text{BCs}}}{\partial \wout } = \sum_{\mu=L,R} 2H_\mu^T\left(H_\mu\wout - B_\mu(t)\right),
\end{equation}
where $H_\mu = H(t,\mu)$. Piecing this all together yields:
\begin{align}
    \wout =& \left(\hat{D}_H^T\hat{D}_H +  \sum_{\mu=0,L,R}H_\mu^TH_\mu\right)^{-1} \nonumber \\
    &\left(\hat{D}_H^Tf(t,x) + \sum_{\mu=0,L,R}H_\mu^TQ_\mu(t,x)\right),
\end{align}

where $Q_0 =g(x)$, $Q_L=B_L(t)$, and $Q_R=B_R(t)$.

We therefore show it is equally feasible to obtain an analytic set of linear weights to determine solutions to PDEs. Indeed, the accuracy of the solution depends heavily on how well the hidden states $H$ span the solution space.
To encourage a representative hidden space, we typically bundle train a single network on different equations, namely a network with multiple outputs.

\section{Results}

We investigate our method on numerous well known differential equations of practical interest. We present a summary of our results in Table \ref{table:results}. The full training regime, network design and test conditions can be found in the appendix. For ODEs we report accuracy as the MSE of the residual: $|\hat{D}_n\phi_{\theta}(t)-f(t)|^2$. For PDEs we report accuracy as the MSE between the predicted solution and the analytic solution: $|\psi_{gt} -\psi_{pred}|^2$.

In addition, we highlight how our solver can be used to tackle specific challenges with learning from these equations. 

\subsection{ODEs}
\subsubsection{Linear ODEs}
\begin{figure}[ht!]
    \centering
    \hspace*{-.8cm}
    \includegraphics[width=.45\textwidth]{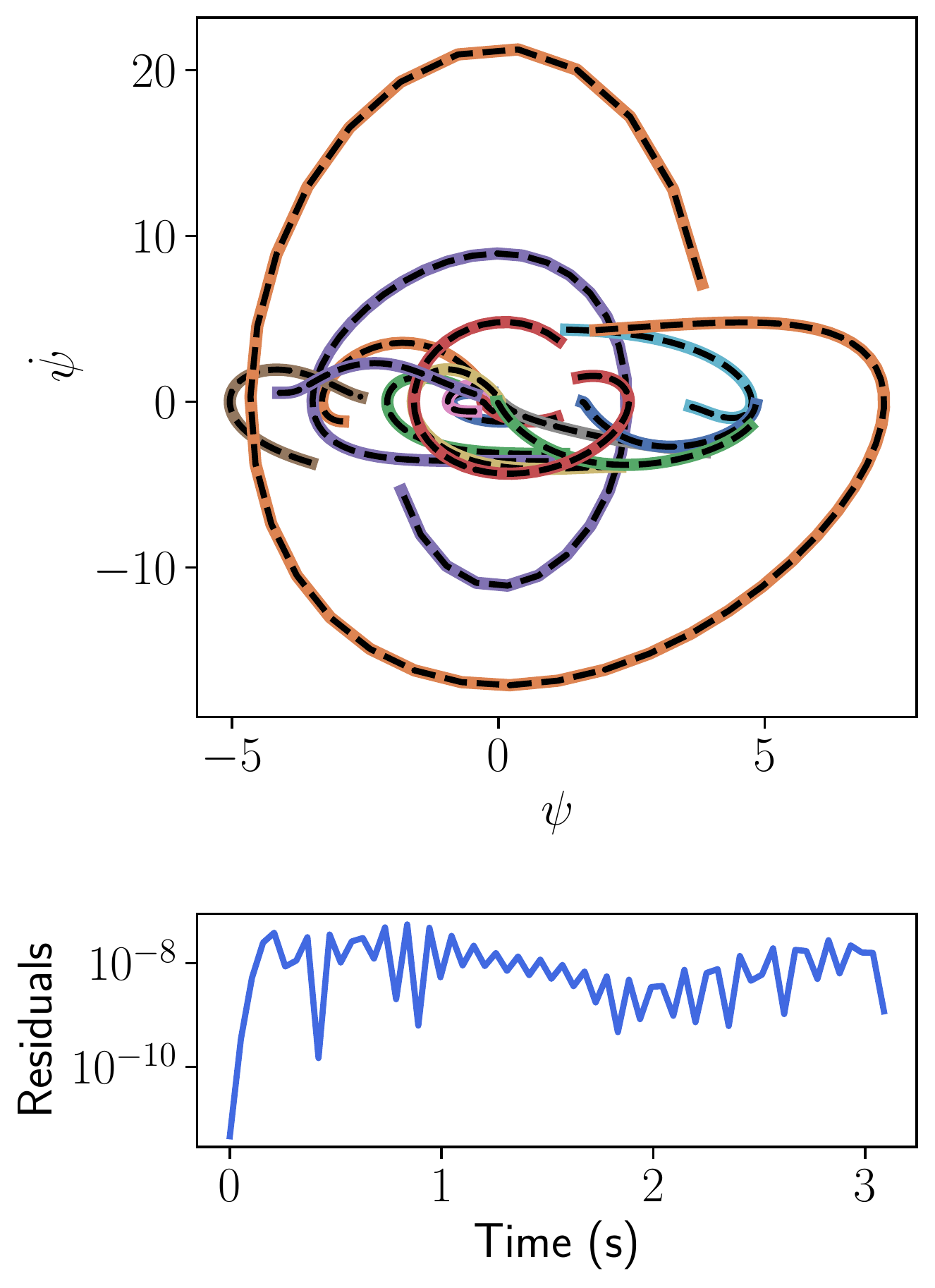}
    \caption{Predicted (colored) versus ground truth (dashed black) phase space, namely a plot of  space against velocity for different times, to 20 second-order non-homogeneous ordinary differential equations. Average residuals are shown in the bottom panel.}
    \label{fig:second}
\end{figure}

To test the performance of our approach, we train both first- and second-order methods of linear non-homogeneous differential equations. For first order ODEs, the  equation is defined by the operator of Eqn. \ref{eqn.ode2} with $n=1$ and by specifying three quantities: the time-dependent coefficients $a_0$, the forces $f$, and the  ICs. As such, any first-order linear non-homogeneous ODE can be defined by the tuple $(a_0,f,\text{ICs})$. Given a pre-defined list of options for each quantity in the tuple (see Appendix), we randomly sample 10 tuples for training. We batch train our model on all the equations simultaneously (i.e. $\wout \in \mathbb{R}^{t\times 10}$). We then carry out inference on 1000 randomly sampled test tuples using analytic $\wout$ from Eqn. \ref{eqn.wout_analytic}. Results are presented in Table ~\ref{table:results}. It is clear that for first-order differential equations, transfer learning analytic $W_{\text{out}}$ is significantly advantageous since we obtain high-fidelity solutions. As such, we take a similar approach for second order differential equations described by the operator of Eqn. \ref{eqn.ode2} for $n=2$. We plot the results of 20 ODEs from the test set and compute their residuals in Fig.~\ref{fig:second} where $\dot \psi = d\psi/dt$. Indeed, the overall test accuracy depends on how well the hidden states span the space of differential equations. We typically find that more training bundles results in better test performance (see Appendix). 

\subsubsection{Systems of ODEs}

\begin{figure}[ht!]
    \centering
    \hspace*{-1cm}       
    \includegraphics[width=.434\textwidth]{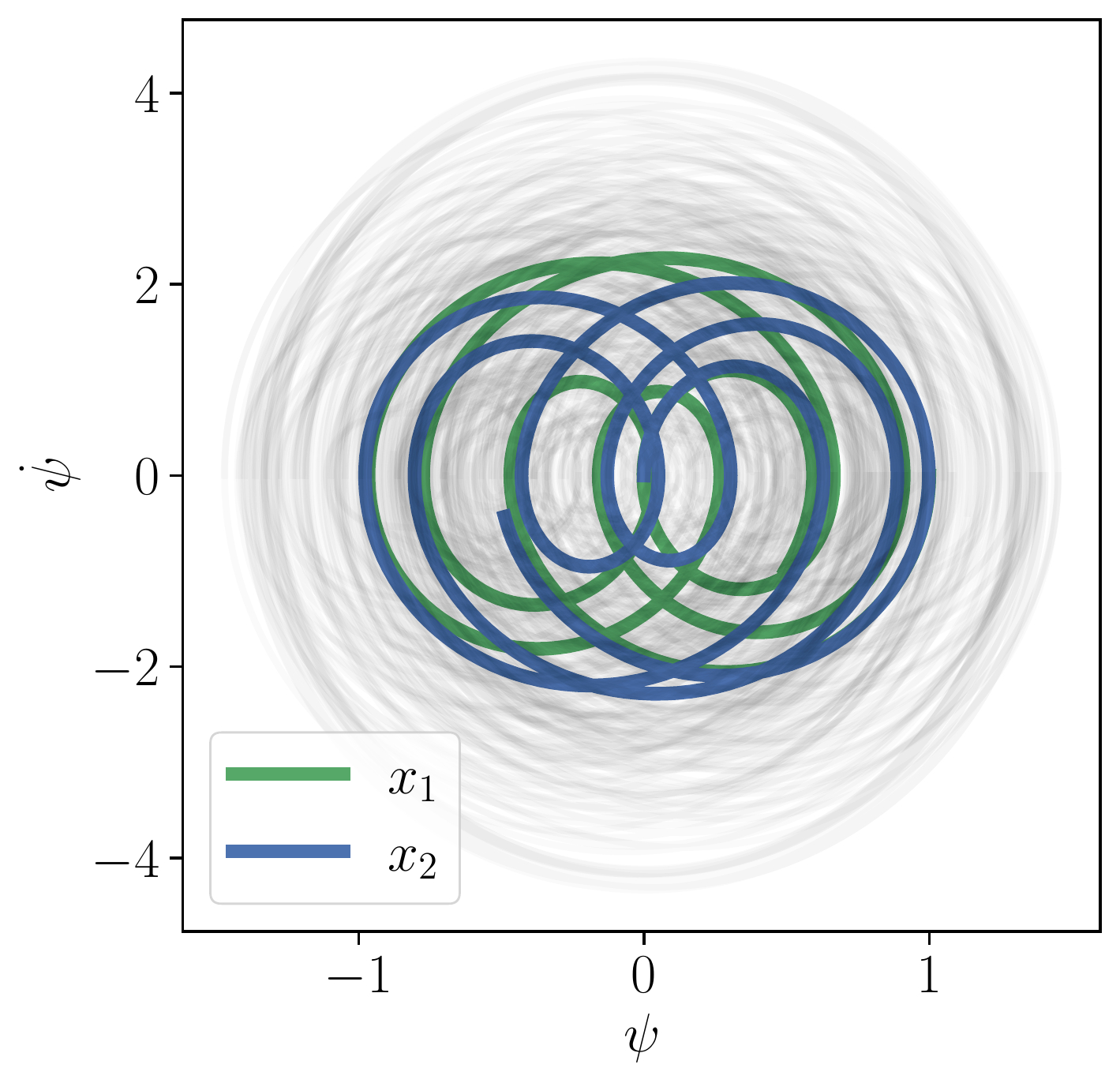}
    \hspace*{-1cm}       
    \includegraphics[width=.45\textwidth]{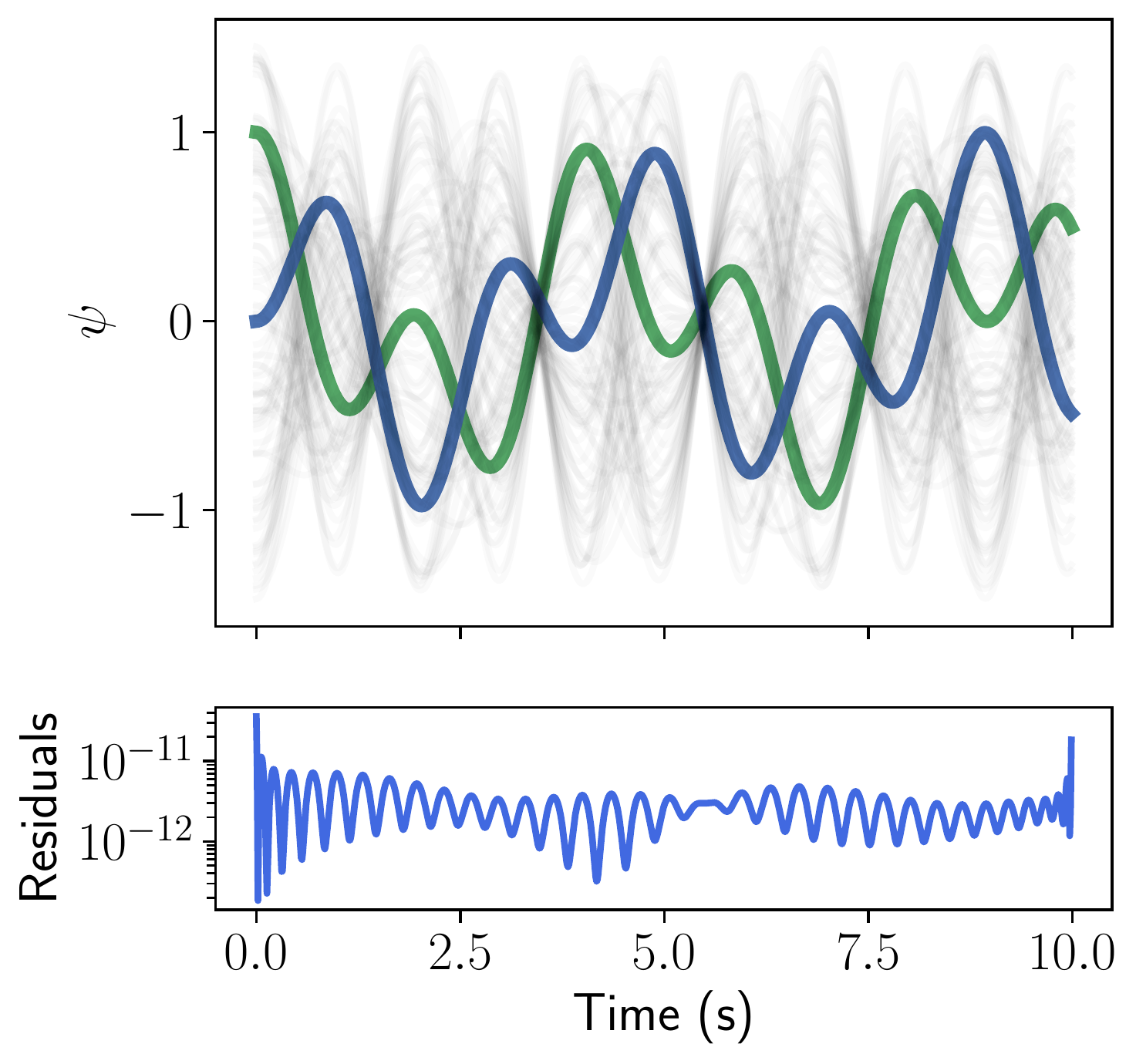}
    \caption{Phase space trajectories of the coupled oscillator system for fixed mass and spring constants (top) and spatial  solutions  (middle). One solution that induces beats is highlighted in color while the other solutions appear in grey. The average residuals of the total realizations are shown in the bottom panel. The initial state of the masses influences how close the normal mode frequencies get. Our network can identify solutions to all 100 initial conditions in $\sim 10^{-2}$ seconds.}
    \label{fig:beats}
\end{figure}

Since an analytic $W_{\text{out}}$ can be computed for linear ODEs, the method naturally extends to systems of linear differential equations (see Appendix for derivation). To highlight this, we investigate a system  of linear second-order ODEs of the form $\ddot{\boldsymbol{\usol}} = A\boldsymbol{\usol}$ that describe a system of two coupled oscillators where $\boldsymbol{\usol} = [\usol_1,\usol_2]$ and $A$ describes the coupling. The equation is of the form:
\begin{equation}
    \begin{bmatrix}
    m & 0 \\
    0 & m 
    \end{bmatrix}
    \begin{bmatrix}
    \ddot{\usol}_1\\
    \ddot{\usol}_2
    \end{bmatrix}= \begin{bmatrix}
    k_1 + k_2 & - k_2\\
    -k_2 & k_1 + k_2
    \end{bmatrix}
    \begin{bmatrix}
    \usol_1 \\
    \usol_2
    \end{bmatrix}.
\end{equation}
We train the network to satisfy 10 different $\{m,k_1,k_2\}$ values and initial conditions $[\boldsymbol{\usol}_0,\boldsymbol{\dot{\usol}}_0]$ (see Appendix for sampling details) such that the pre-trained network can be used to instantaneously compute accurate solutions for different ICs of the coupled masses. We report the result of testing 100 different systems sampled from the same range as training in Table \ref{table:results}. Furthermore, we investigate an interesting application in which the network can be exploited to identify initial conditions of a coupled-oscillator system capable of inducing \textit{beats} - when two normal mode frequencies come close (see Fig. \ref{fig:beats}) \cite{schwartz_lecture_2017}.

\subsubsection{Nonlinear ODE}
\begin{figure}[ht!]
    \centering
    \includegraphics[width=0.45\textwidth]{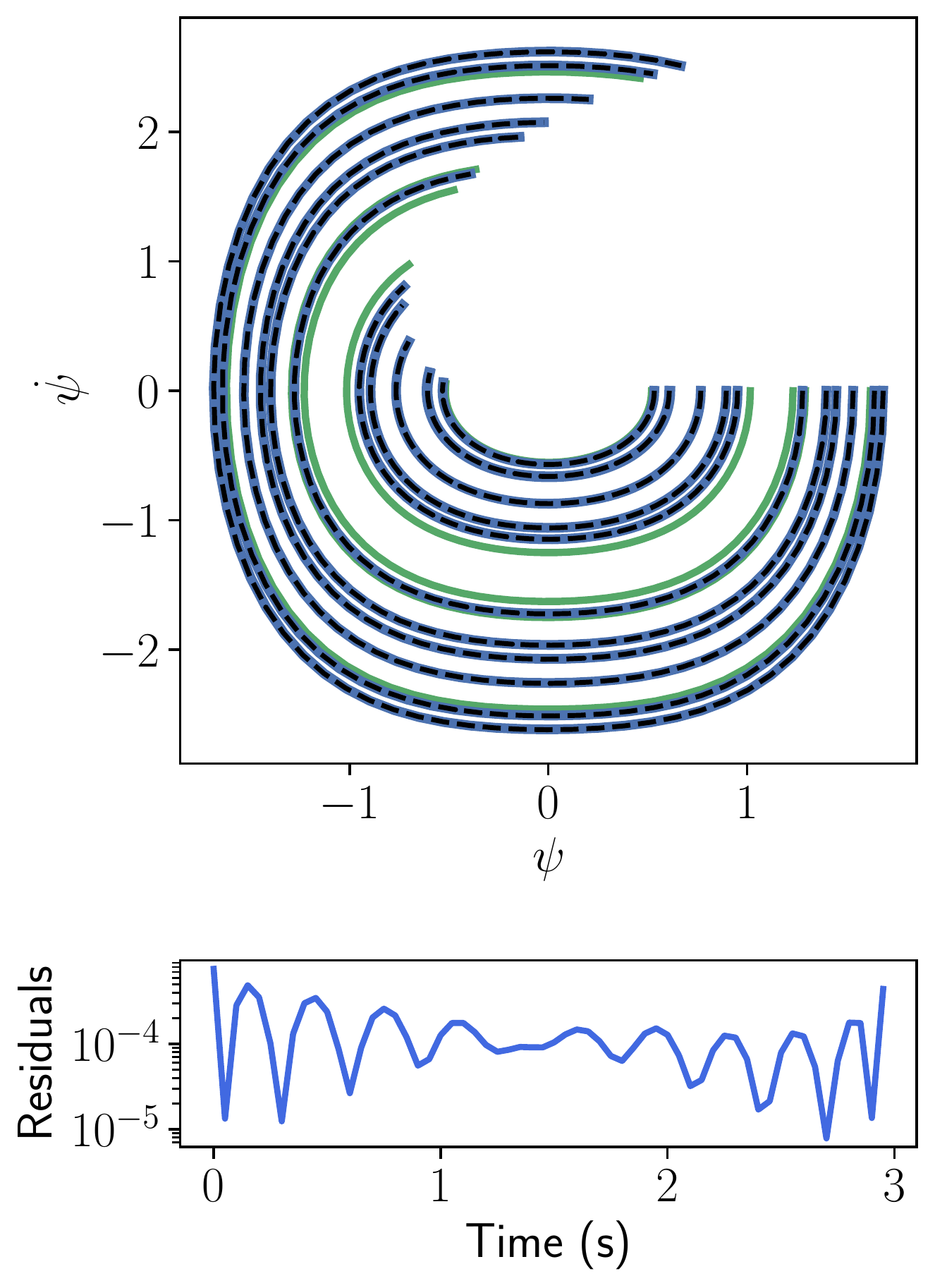}
    \caption{Top: phase space of predicted trajectories of a nonlinear oscillator system. The training curves ares shown in green and the test in blue. Dashed black lines represent ground truth solutions. Bottom: average residuals of 30 predicted solutions across different initial conditions.}
    \label{fig:nlosc}
\end{figure}

Until this point, we have only investigated the success of fine-tuning $W_{\text{out}}$ analytically for linear ODEs, nevertheless our network proposal can still be exploited to transfer learn nonlinear ODEs. To do so, we  replace the computation of analytic $W_{\text{out}}$ with gradient-based optimization. Note, since the hidden weights are frozen, the final hidden activations can be pre-computed given sampling points $t$ for efficient optimization. 
We use this approach to solve a Hamiltonian nonlinear system  described by the ODE:
\begin{equation}
    \ddot{\usol} = -\usol - \usol^3    ,
\end{equation}
that conserves energy given by the Hamiltonian: 
\begin{equation}
    \mathcal{H} = \frac{\dot{\usol}^2}{2} + \frac{\usol^2}{2} + \frac{\usol^4}{4}.
\end{equation}
We train the system on 5 initial positions $\psi_0$ randomly sampled in the range [0.5, 2.0] and with initial velocity $\dot \psi_0=0$. The loss function during training consists of (1) a differential equation loss, (2) an initial condition loss, and (3) an energy conservation loss penalty. The energy loss enforces the Hamiltonian  at all  points in time to be the same, namely $\mathcal{L}_\text{E} = (\mathcal{H}(\usol,\dot{\usol}) - \mathcal{H}(\usol_0,\dot{\usol}_0))^2$ \cite{mattheakis_hamiltonian_2020}.
We then evaluate the performance of the hidden states on 30 ICs sampled in the same range (see Fig.~\ref{fig:nlosc}). Since we freeze the hidden layers, we can pre-compute the hidden activations $H(\bar{t})$ at fixed time $\bar{t}$ and then fine-tune $W_{\text{out}}$ using gradient descent for 5000 epochs. Note that the optimization can be done using other methods as well, including L-BFGS since the entire problem is reduced to convex optimization.

\subsection{PDEs}
PDEs can be used to model complex spatio-temporal systems making them of practical interest in numerous domains. Typically, the most well-studied PDEs are linear and include the diffusion, Poisson, and wave equations. To benchmark the performance of this approach, we investigate the Poisson equation and the time-dependent Schr\"{o}dinger equation.

\subsubsection{Poisson Equation}

\begin{figure}[ht!]
    \centering
    \includegraphics[width=0.45\textwidth]{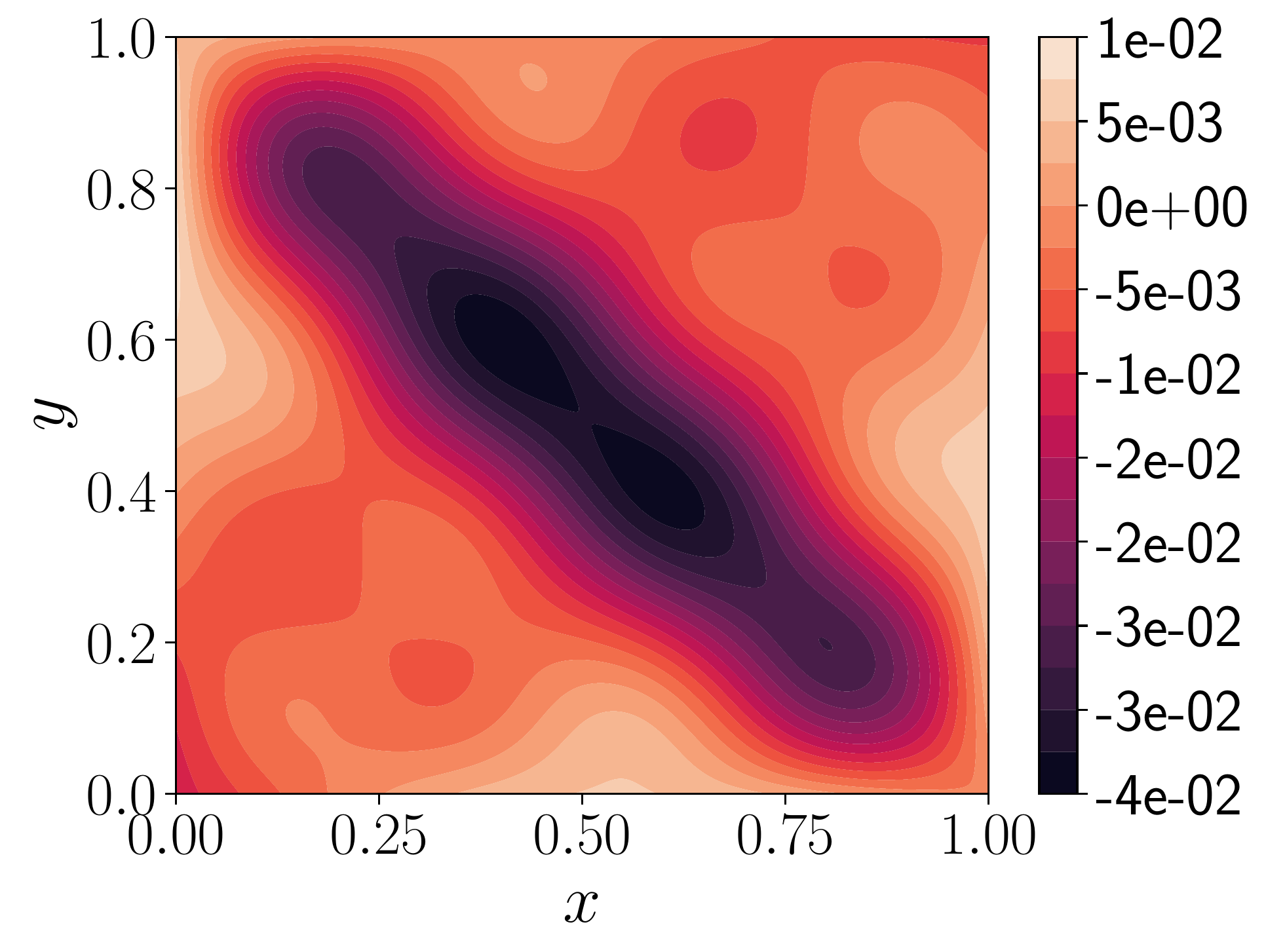}
    \includegraphics[width=0.45\textwidth]{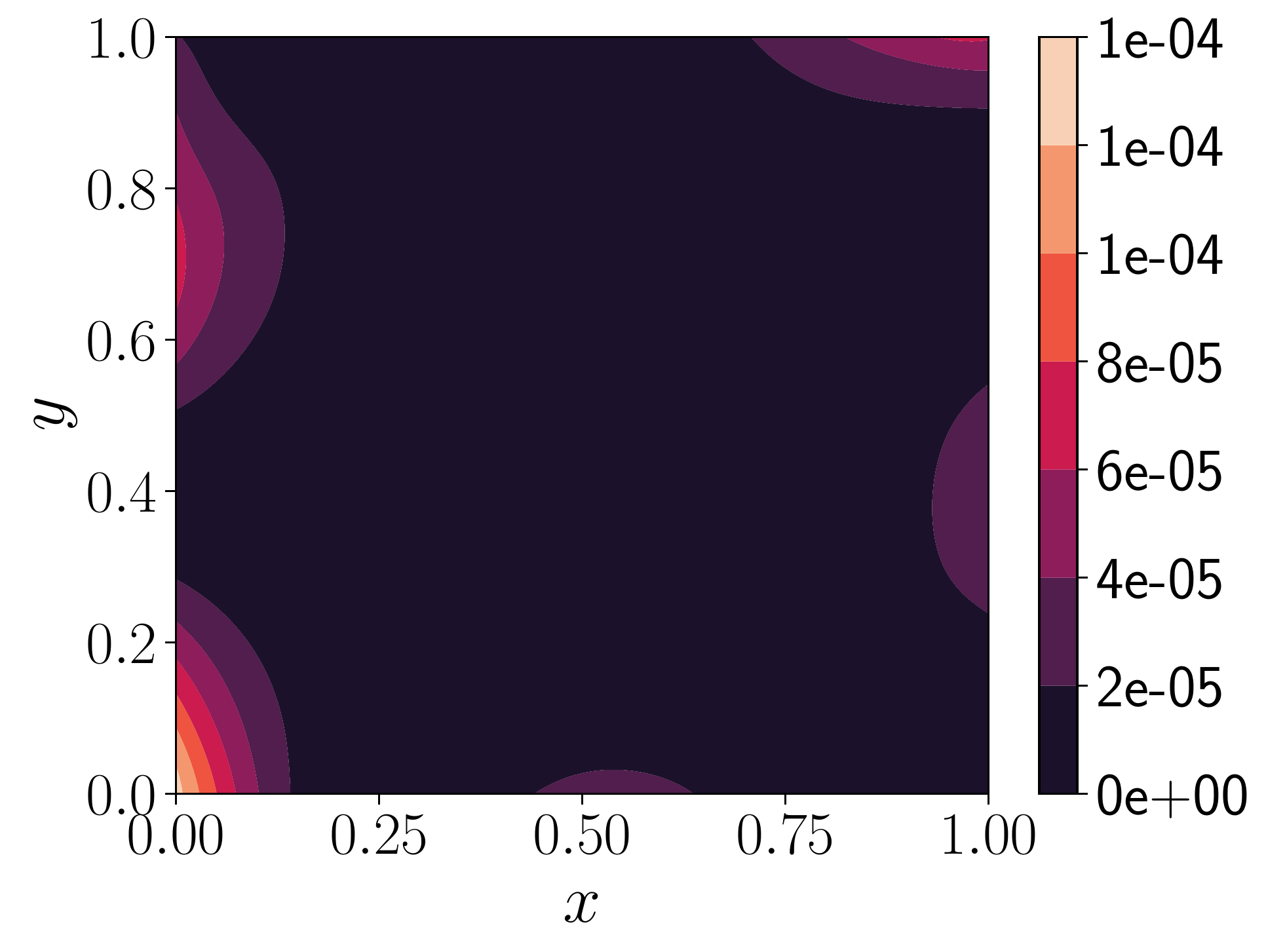}
    \caption{Predicted solution (top) of the Poisson equation with an  initial charge distribution $\rho(x,y)$ composed of multiple frequencies $k$. The network is pre-trained on the individual frequencies and can obtain the solution to the combination in one-shot (35s) with high fidelity/low MSE (bottom).
    }
    \label{fig:poisson}
\end{figure}

The Poisson equation is an extensively studied PDE in physics, typically used to identify an electrostatic potential $\psi$ given a charge distribution $\rho$. In 2-D, it can be described by:
\begin{equation}
 \frac{\partial^2\usol}{\partial x^2} +  \frac{\partial^2\usol}{\partial y^2} = \rho(x,y).
\end{equation}
We  define this PDEs in the domain $x\in[x_L,x_R]$ and $y\in[y_B,y_T]$ with BCs:
\begin{equation}
    \usol(x_L,y) =  \usol(x_R,y) = 
    \usol(x,y_B) =  \usol(x,y_T) = 0.
\end{equation}
We train the network on 4 different charge distributions $\rho(x,y) = {\sin(k\pi x) \sin(k\pi y)}$ for $k \in {1,2,3,4}$. We then evaluate the performance of our network in two settings. The first is an ablation across 100 linearly spaced values of $k$ in $[1,4]$ (see Table \ref{table:results}). As a second experiment we test the proposed transfer learning method  on a harder testing force function  of the form: 
\begin{equation}
\rho_{\text{test}} = \frac{1}{4}\sum_{k=1}^4(-1)^{k+1}2k \sin(k\pi x)\sin(k\pi y).    
\end{equation}
The solution is shown in the top graph of Fig.~\ref{fig:poisson}. To assess the network performance, we plot in the lower graph of Fig.~\ref{fig:poisson} the mean square error (MSE) computed between  the predicted $\psi(x,y)$ and the  analytical solution  which reads:
\begin{equation}
    \usol(x,y) = \frac{-1}{2(k\pi)^2} \sin(k\pi x) \sin(k \pi y).
\end{equation}

\subsubsection{Time-Dependent Schr\"{o}dinger Equation}

In quantum mechanics the time-dependent Schr\"{o}dinger equation describes the propagation of a  wavefunction  through space and time. The PDE in  one-dimensional space is of the form:
\begin{equation}
   i\hbar \frac{\partial }{\partial t}\psi(x,t) =\left[ -\frac{\hbar^2}{2m} \frac{\partial^2}{\partial x^2}  + V(x) \right] \psi(x,t),    
\label{se}
\end{equation}
where $\psi({x},t)$ is a complex-valued function called  wave-function,  $V(x)$ is a stationary potential function, $m$ is the mass, and $\hbar$ is a constant.
We investigate the  quantum free-particle evolution for which $V(x) = 0$ for several initial states $\psi(x,0)$. 

\begin{figure}[t!]
    \centering
    \includegraphics[width=.45\textwidth]{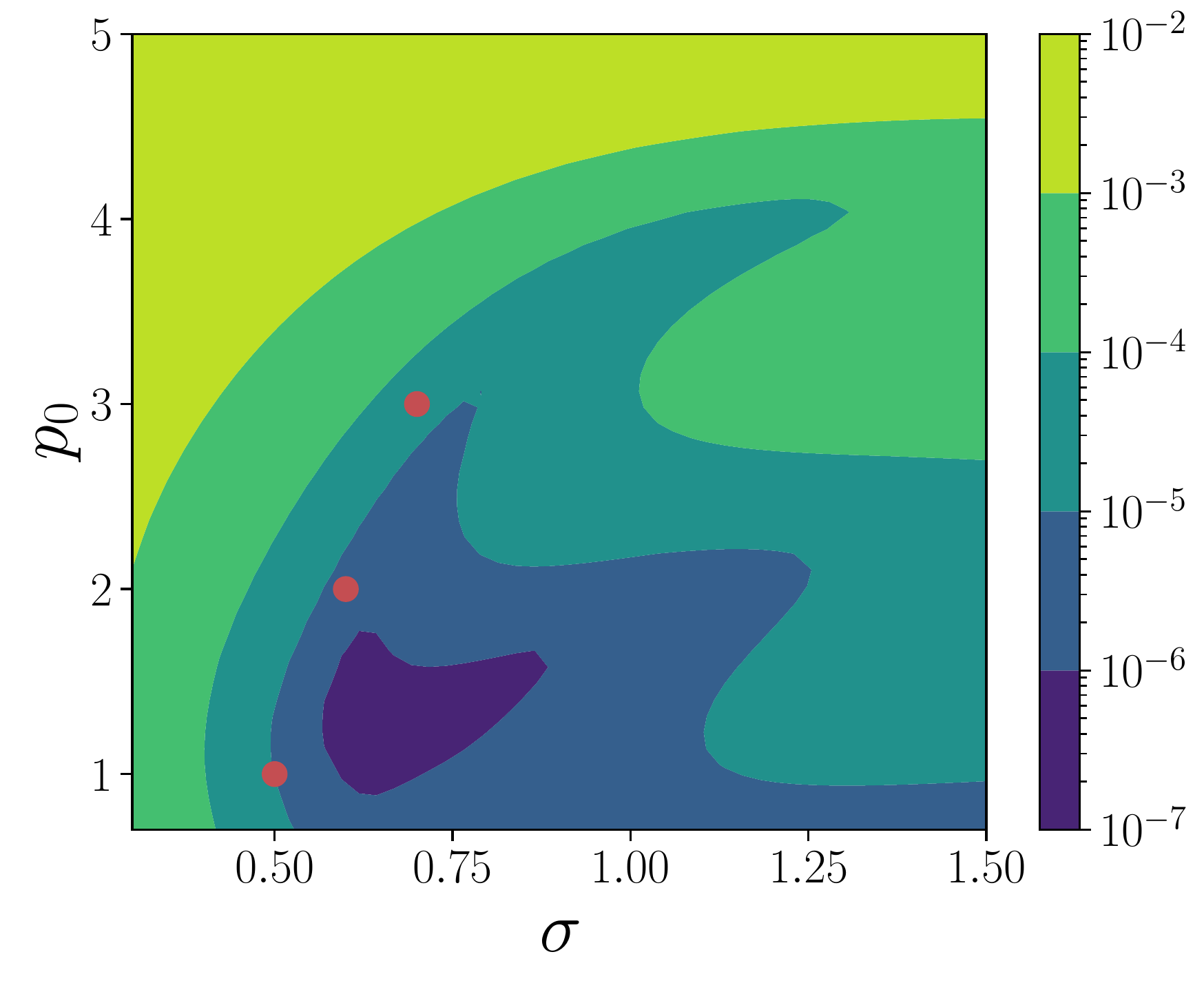}
\caption{MSE between predicted and analytic solutions $|\psi|^2$ as a function of $\sigma$ and $p_0$. Red circles represent the three configurations for which the network was batch trained. We see that as $p_0$ increases, transfer learning $\wout$ becomes less effective because of the F-principle bottleneck for PINNs.}
\label{fig:schroed}
\end{figure}

To train the complex-valued wave-function, we separate the real $\psi_R$ and imaginary $\psi_I$  parts  \cite{raissi_physics-informed_2019}, namely $ \psi(x,t) = \psi_R(x,t) + i\psi_I(x,t)$. 
%
By plugging the above form of $\psi$ into Eqn. \ref{se} we obtain a coupled system of real-valued PDEs as:
\begin{equation}
    \frac{\partial }{\partial t} 
    \begin{bmatrix}
    \psi_R \\
    \psi_I
    \end{bmatrix}
    =
    \begin{bmatrix}
    0 & -\hbar/2m\\
    \hbar/2m & 0
    \end{bmatrix}
    \frac{\partial^2 }{\partial x^2} 
    \begin{bmatrix}
    \psi_R\\ \psi_I 
    \end{bmatrix},
\label{coupled_pde}
\end{equation}
which is of the form $\boldsymbol{\usol}_t = A\boldsymbol{\usol}_{xx}$.
The system is linear and thus, we can obtain  analytic $\wout$. We consider a network with two outputs per equation associating with $\psi_R$ and $\psi_I$, where each output is, respectively, described by a set of weights as $\wout=\left[ W_\text{R}, W_\text{I}\right]^T$. Then,  the network solutions read:

\begin{equation}
    \begin{bmatrix}
    \psi_R\\
    \psi_I
    \end{bmatrix}
    = 
    \begin{bmatrix}
    H & 0\\
    0 & H
    \end{bmatrix}
    \begin{bmatrix}
    W_R\\
    W_I
    \end{bmatrix},
\end{equation}
By taking the L$_2$ loss of  Eqn. \ref{coupled_pde} as well as the BCs and ICs we obtain:
\begin{equation}
\begin{split}
    \wout =& (D_H^T D_H + H_0^T H_0 \\
    &+ H_d^TH_d + \dot{H}_d^T\dot{H}_d)^{-1}(H_0^T\psi_0),
\end{split}
\end{equation}
where $H_0=H(0,x)$, $H_d=H(t,L)-H(t,R)$, and $\dot{H}_d = H_x(t,L) - H_x(t,R)$.

To investigate a particular set of solutions, we define the initial condition for this problem as:
\begin{equation}
    \psi(x,0)= \frac{1}{\pi^{1/4}\sqrt{\sigma}}e^{-{(x-x_0)^2}/{(2\sigma^2)}+ip_0x/\hbar},
\end{equation}
 that leads to the exact solution:
\begin{equation}
    \psi(x,t) = \frac{e^{-\frac{(x-(x_0+p_0t/m))^2}{2\sigma^2(1+i\hbar t/m\sigma^2)}}e^{i(p_0x-Et)/\hbar}}{\pi^{1/4}\sqrt{\sigma(1+i\hbar t/m\sigma^2)}},
    \label{eqn.exactFreeParticle}
\end{equation}
where $E = p_0^2/2m$.

\begin{figure}[t!]
    \centering
    \includegraphics[width=.48\textwidth]{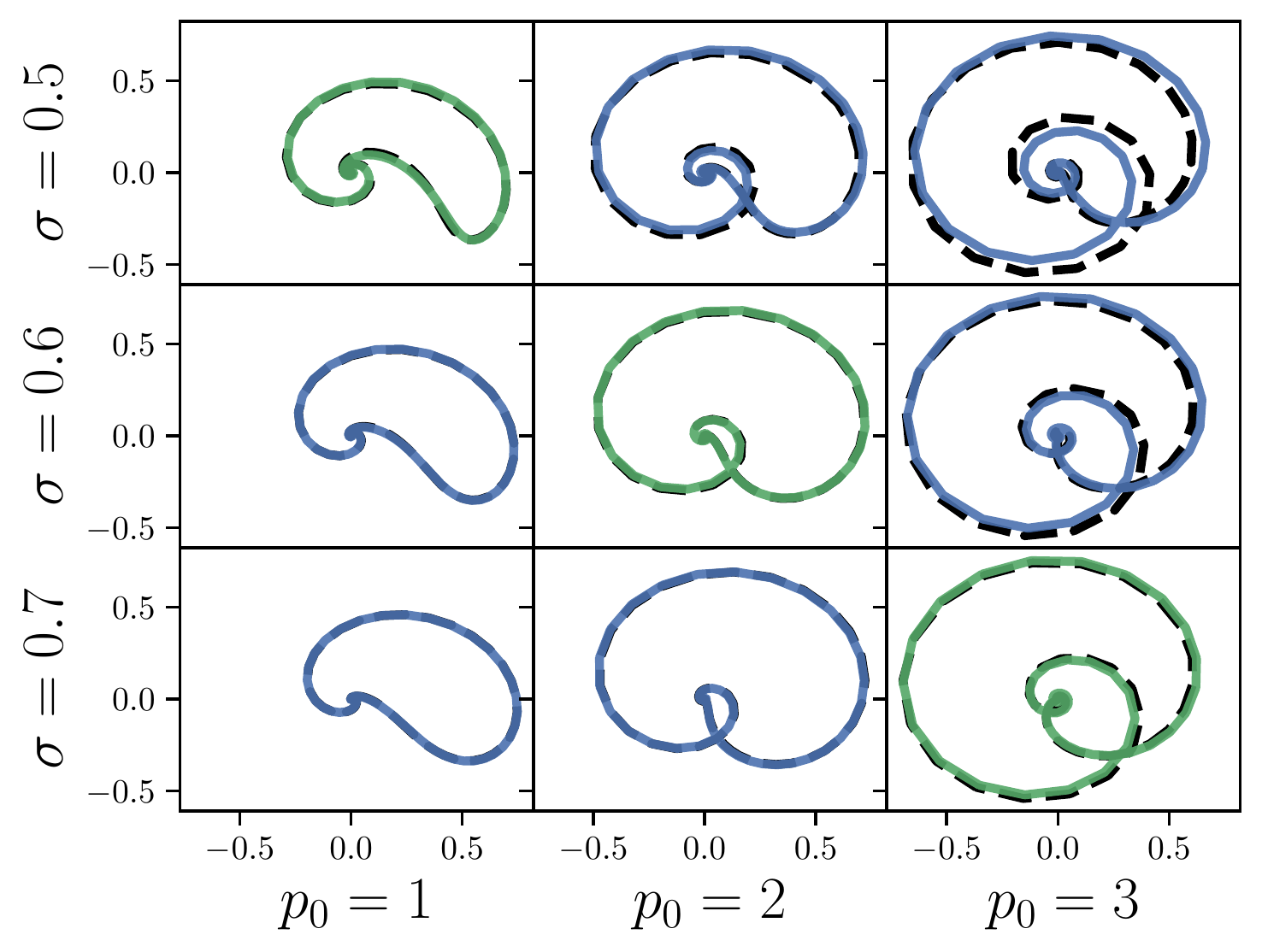}
\caption{Real (x-axis) against imaginary (y-axis) wave-functions  of the predicted solutions $\psi(T_{\max},x)$ for different realizations of $\sigma,p_0$ with solid and dashed lines representing the predicted and ground truth solutions. The diagonal configurations are used for training. Non-diagonals constitute test configurations.}
\label{fig:schroed2}
\end{figure}

We train a network simultaneously on three solutions of Eqn. (\ref{se}) with pairs of $\sigma,p_0$ such that the network  is trained for $(\sigma,p_0)=\{ (0.5,1),(0.6,2),(0.6,3) \}$. We show that by using only 3 training ICs, accurate solutions to multiple other configurations can be obtained instantly and with high accuracy (see Fig. \ref{fig:schroed}). 
We measure and present in  Fig. \ref{fig:schroed} the accuracy of our predicted solution by computing the MSE across time and space against the analytic solution Eqn. \ref{eqn.exactFreeParticle}.   
Furthermore, we show that near the bundles, the predicted real-imaginary complex space under different $\sigma,p_0$ pairs tightly couples to the analytic solution (see Fig. \ref{fig:schroed2}). Our results also highlight the F-principle, a conclusion drawn about deep networks which shows that high frequency components require more training \cite{xu_training_2019} as can be seen in Fig. \ref{fig:schroed} by looking at higher $p_0$ values which induce higher frequency components in the solution $\psi$.

\section{Conclusion}

We have extensively shown how PINNs can be batch trained on a family of differential equations to learn a rich latent space that can be exploited for transfer learning. For linear systems of ODEs and PDEs, the transfer can be reduced to computing a closed-form solution for $W_{\text{out}}$ resulting in one-shot inference. This analytic solution significantly speeds up  inference on unseen differential equations by orders of magnitude, and can therefore replace or augment traditional transfer learning. In particular, we show that such a network can identify first-order ODEs, second-order coupled ODEs, Poisson and Schr\"{o}dinger equations with high levels of accuracy within a few seconds. Furthermore, in the nonlinear setting, where a closed-form analytic $W_{\text{out}}$ is not  derived, we show that our approach can still be used with gradient descent to identify accurate solutions to the nonlinear oscillator system. These results are particularly important to practitioners who seek to rapidly identify accurate solutions to  differential equations  of the same type, namely of the same order, but under different conditions and coefficients.
Indeed many new applications may arise as a consequence of this approach, from transfer learning on large domains, to solving high dimensional linear PDEs. Future work in this direction may  adapt to  data-dependent settings, incorporate non-linear outputs to develop one-shot training for nonlinear equations, and investigate properties of the learnt hidden space.
\bibliographystyle{icml2022}
\bibliography{references.bib}

\newpage
\appendix
\onecolumn

\section{System of second order differential equations}
To compute the analytic $\wout$ for a system of second order differential equation we begin by defining:

$$
 \ddot {\boldsymbol{\usol} }= A\boldsymbol{\usol}
$$

where dots denote time derivatives and 
$$ \boldsymbol{\usol} = \begin{bmatrix}
H & 0 \\
0 & H \\
\end{bmatrix} 
\begin{bmatrix}
W_q\\
W_p
\end{bmatrix}.
$$

Then, by computing the $L2$ loss on the equation above including initial conditions $\usol_0=\usol(0), \dot \usol_0=\dot{\usol}(0)$ we obtain:

$$\wout  =\left(D_H^TD_H + H_0^TH_0 +H_{0d}^TH_{0d}\right)^{-1} \left(H_0^T \usol_0 + H_{0d}^T\dot{\usol}_{0}\right)$$

where
\begin{align*}
D_H  &= \begin{bmatrix}
\ddot{H} & 0 \\
0 & \ddot{H} \\
\end{bmatrix},
\\
H_0  &= \begin{bmatrix}
H(0) & 0 \\
0 & H(0) \\
\end{bmatrix},
\\
H_{0d}  &= \begin{bmatrix}
\dot{H}(0) & 0 \\
0 & \dot{H}(0) \\
\end{bmatrix}.
    \end{align*}
    
\section{QR Decomposition}

In the main manuscript, we define the loss function for ODEs as:
\begin{equation}
    \mathcal{L} = (\hat{D}_n u_{\theta}(t) - f(t))^2 + (\bar{D}_0u_{\theta}(t)- u_{\text{ic}})^2,
    \label{eqn.loss3}
 \end{equation}
 
For ease of notation, let $\hat{D}_n u_{\theta}(t) = \hat{Y}$, $f(t)=Y$, $\bar{D}_0u_{\theta}(t)=\hat{Y}_0$ and $u_{\text{ic}}=Y_0$. The loss function above can be re-written as a single loss function s.t.:

\begin{equation}
    \mathcal{L} = \left(\begin{bmatrix}
    \hat{Y} \\
    \hat{Y}_0
    \end{bmatrix}-
    \begin{bmatrix}
    Y \\
    Y_0
    \end{bmatrix} \right)^2.
\end{equation}

To see this, we can expand the vector notation as:
\begin{equation}
    \mathcal{L} = 
    \left(\begin{bmatrix}
    \hat{Y} & \hat{Y}_0\\
    \end{bmatrix}-
    \begin{bmatrix}
    Y & Y_0 \\
    \end{bmatrix} \right)
    \left(\begin{bmatrix}
    \hat{Y} \\
    \hat{Y}_0
    \end{bmatrix}-
    \begin{bmatrix}
    Y \\
    Y_0
    \end{bmatrix} \right),
\end{equation}
which when expanded resolves to Eqn.\ref{eqn.loss3}.

By differentiating the above loss equation with respect to $W_{\text{out}}$ and setting it to zero, we obtain a linear least squares problem as
\begin{equation}
    \left( H^TH + H_0^TH_0 \right)\wout = \left(H^TY +H^TY_0\right),
\end{equation}
which is of the form
\begin{equation}
    A^TA ~\wout = A^TY.
\end{equation}
Since the left hand-side is a square matrix $A^TA$, it is possible to take its pseudo-inverse. However, for a number of problems, it is possible that the matrix $A$ has a large condition number and therefore the squaring procedure of the normal equations squares the condition number making it unstable. Although it is possible to take a pseudo-inverse, in such cases, rather than using the normal equations to obtain a solution to $\wout$, it is possible to solve for $\wout$ using QR decomposition. In other words:

\begin{equation}
    AW_{\text{out}} = Y,
\end{equation}
and since $A$ is not square, we can decompose it into $A=QR$ such as
\begin{equation}
    W_{\text{out}} = R^{-1}Q^TY.
\end{equation}
One advantage of using QR is that it avoids forming the gram matrix $A^TA$ of the normal equations which can be singular.

\section{Computational Complexity}

One special outcome of our formalism is that it separates the homogeneous part of the differential equation from the initial conditions and forces. In other words, if we investigate the normal equations for $\wout$, we see that
\begin{equation}
    W_{\text{out}} = (\hat{D}_H^T\hat{D}_H + \bar{D}_H^T\bar{D}_H)^{-1}(D_H^Tf'(t) + \bar{D}_H^Tu_{\text{ic}}').
    \label{eqn.wout_analytic_appendix}
\end{equation}

Notice that the forces and initial conditions appear outside the matrix inversion. This is a particularly important feature as it allows us to scale rapidly when the differential equation is fixed and the solution to many initial conditions or forces is required. In fact, the computational complexity of the inversion is $\mathcal{O}(h^3)$ where $h$ is the number of output neurons of the final hidden layer and the multiplication is $\mathcal{O}(h^2m)$, where m is the number of initial conditions and forces. Therefore, if the differential equation is fixed the total computational cost of computing $\wout$ is $\mathcal{O}(h^3 + mh^2)$. However, if the differential equation is not fixed and varies across all m samples, then the inversion has to be computed $m$ times such that  the total computational cost is $\mathcal{O}(mh^3 + mh^2)$.

\section{Training Configuration}
All models are trained using an Adam optimizer with a learning rate of $10^{-3}$. The networks are all trained on a Macbook Pro, 2.2 GHz Intel Core i7, 16 Gb RAM. We use 64-bit tensors. 

\begin{table}[htbp]
 \resizebox{\textwidth}{!}{%
\begin{tabular}{ccccccccc}
\textbf{Differential Equation} & \textbf{Architecture (N bundles)} & \textbf{\# training Bundles} & \textbf{Training Iterations} & \textbf{\# Training Collocation Points} & \textbf{Evaluation Domain} & \textbf{Evaluation Deltas} & \textbf{Activations} & \textbf{Optimizer} \\
First-Order Linear Inhomogeneous   & 1-100-100-1*N & 10    & 10000 & 30    & t in [0,3] & dt = 0.1 & tanh  & Adam \\
Second-Order Linear Inhomogeneous   & 1-100-100-1*N & 10    & 10000 & 30    & t in [0,3] & dt = 0.05 & tanh  & Adam \\
Coupled-Oscillator  & 1-100-100-2*N & 10    & 10000 & 50    & t in [0,10] & dt = 0.01 & sin   & Adam \\
Non-Linear Oscillator   & 1-100-100-1*N & 5     & 10000 & 60    & t in [0,3] & dt = 0.05 & sin   & Adam \\
Poisson    & 2-100-100-1*N & 4     & 40000 & 1000  & x in [0,1], t in [0,1] & dx = 0.01, dt = 0.01 & sin   & Adam \\
Schroedinger  & 2-100-100-2*N & 3     & 40000 & 1000  & x in [-10,10], t in [0,1] & dx = 0.1, dt = 0.01 & $\alpha \sin + (1-\alpha)\tanh$ & Adam \\
\end{tabular}%
 }%
\end{table}%

\textbf{First-Order Linear Inhomogeneous}

The equation is of the form:
\begin{equation}
    \dot{u} +a(t)u = f(t),
\end{equation}
subjected to an initial condition $u_0$.
We sample within:
\begin{align*}
   f & \in \{ \cos(t), \sin(t),t\}, \\ 
 a & \in \{t,t^2,1\},  \\
  (u_0) & \in [-5,5].
 \end{align*}

\textbf{Second-Order Linear Inhomogeneous}

The equation is of the form:

\begin{equation}
    \ddot{u} +a_1(t)\dot{u} +a(t)u = f(t)
\end{equation}
with initial conditions $u_0$ and $\dot u_0$.
We  sample within:
\begin{align*}
f  &\in \{1,t, \cos(t), \sin(t)\}, \\
a & \in \{1,3t,t^2\}, \\
a_1 & \in \{1,t^2,t^3\}, \\
 (u_0,\dot u_0) & \in [-5,5]. 
\end{align*}

\textbf{Coupled oscillator}

The equation is of the form:

\begin{equation}
    \begin{bmatrix}
    m & 0 \\
    0 & m 
    \end{bmatrix}
    \begin{bmatrix}
    \ddot{u}_1\\
    \ddot{u}_2
    \end{bmatrix}= \begin{bmatrix}
    k_1 + k_2 & - k_2\\
    -k_2 & k_1 + k_2
    \end{bmatrix}
    \begin{bmatrix}
    u_1 \\
    u_2
    \end{bmatrix}.
\end{equation}

We sample:
\begin{align*}
    m &\in [1,2] , \\ 
   (k_1, k_2) & \in [0.5,4.5], \\
 (u_0,u_1,\dot{u}_{1,0},\dot{u}_{2,0}) & \in [-1.5,1.5].
\end{align*}

\textbf{Nonlinear oscillator}

The equation is:
\begin{align}
   \ddot{u} + u + u^3 = 0.
\end{align}

We sample within:
\begin{align*}
    (u_0, \dot{u}_0) \in [0.5,2].
\end{align*}

\textbf{Poisson equation}
\begin{equation}
    \nabla^2 u = \rho
\end{equation}

we sample:
\begin{align*}
    \rho \in \{ \sin(x) \sin(y), \sin(2x) \sin(2y), \sin(3x)\sin(3y), \sin(4x) \sin(4y) \}.
\end{align*}

\textbf{Schrodinger}

\begin{equation}
    \begin{bmatrix}
    \dot{u}_R\\
    \dot{u}_I
    \end{bmatrix}= \begin{bmatrix}
    0 & -\hbar/2m\\
    \hbar/2m & 0
    \end{bmatrix}
    \begin{bmatrix}
    u''_R \\
    u''_I
    \end{bmatrix}
\end{equation}

where $\psi = (u_R, u_I)$ and:

\begin{equation}
    \psi(x,0)= \frac{1}{\pi^{1/4}\sqrt{\sigma}}e^{-{(x-x_0)^2}/{(2\sigma^2)}+ip_0x/\hbar}. 
\end{equation}

and 
\begin{align*}
\sigma &\in \{0.5,0.6,0.7\}, \\ 
p_0 & \in \{1,2,3\} .
\end{align*}


\section{Training Bundles}

Typically, the more training bundles we use, the more diverse the hidden states have to be in order to generate accurate solutions for all the training differential equations. As such, we would expect that a diverse set of hidden states should also result in better inference for unseen differential equations. Experimentally, we find and show in Fig. \ref{fig:SMfigure} that for first order differential equations this is true:

\begin{figure}[h!]
    \centering
    \includegraphics[width=.4\textwidth]{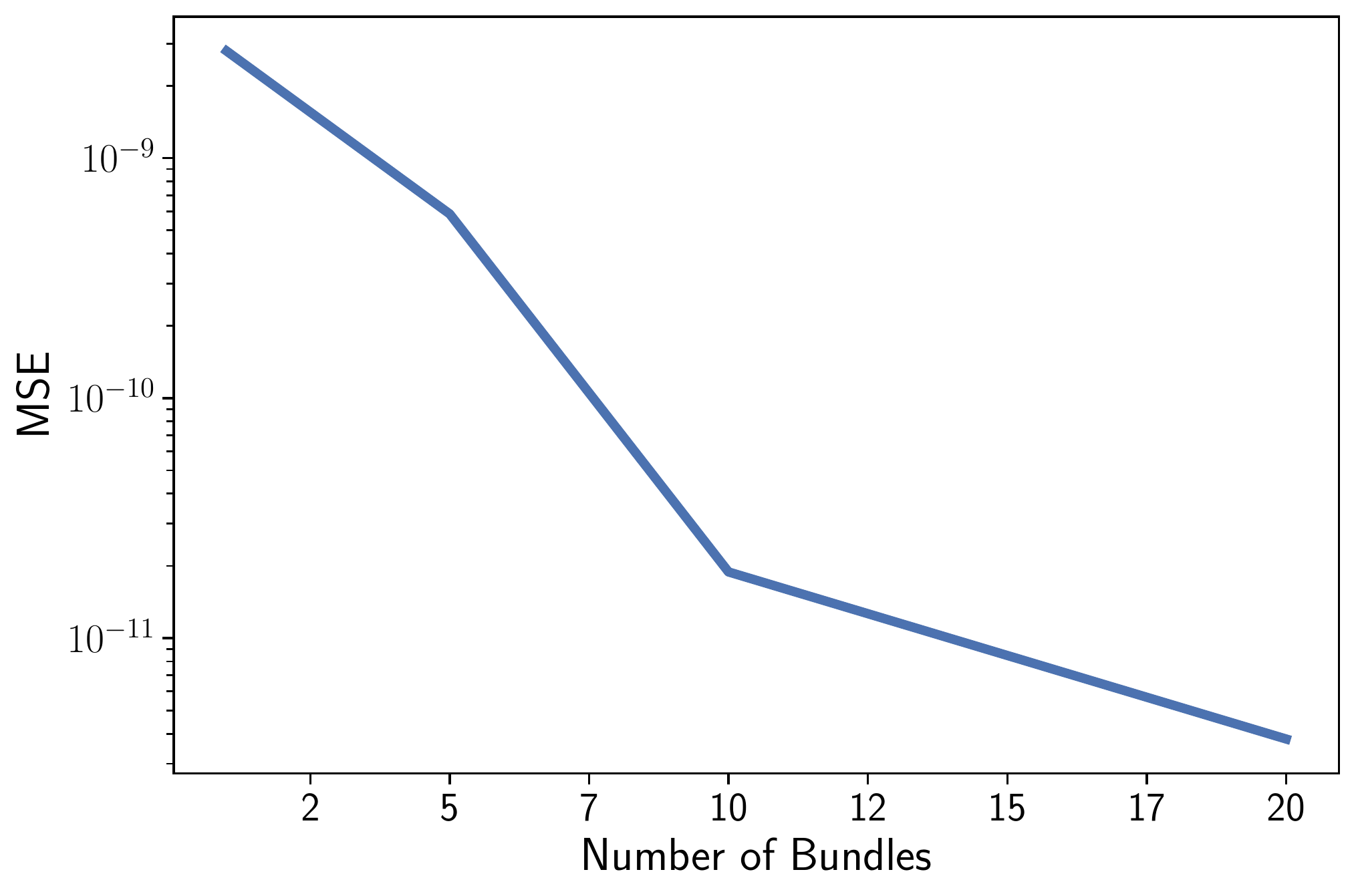}
    \caption{Test MSE as a function of number of bundles. The more bundles we use to train, the better our test accuracy gets.}
    \label{fig:SMfigure}
\end{figure}

Of course doing this also increases the training overhead, thus empirically we use 10 consistently across our experiments.

\end{document}